\documentclass[sigconf]{acmart}
\usepackage{balance} 
\usepackage{subfigure}
\usepackage{hyperref}
\usepackage[capitalise]{cleveref}
\usepackage{adjustbox}
\AtBeginDocument{%
 }

\copyrightyear{2024}
\acmYear{2024}
\setcopyright{rightsretained}
\acmConference[SIGIR '24]{Proceedings of the 47th International ACM SIGIR Conference on Research and Development in Information Retrieval}{July 14--18, 2024}{Washington, DC, USA}
\acmBooktitle{Proceedings of the 47th International ACM SIGIR Conference on Research and Development in Information Retrieval (SIGIR '24), July 14--18, 2024, Washington, DC, USA}
\acmDOI{10.1145/3626772.3657927}
\acmISBN{979-8-4007-0431-4/24/07}

\makeatletter
\gdef\@copyrightpermission{
  \begin{minipage}{0.3\columnwidth}
   \href{https://creativecommons.org/licenses/by/4.0/}{\includegraphics[width=0.90\textwidth]{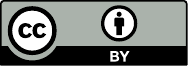}}
  \end{minipage}\hfill
  \begin{minipage}{0.7\columnwidth}
   \href{https://creativecommons.org/licenses/by/4.0/}{This work is licensed under a Creative Commons Attribution International 4.0 License.}
  \end{minipage}
  \vspace{5pt}
}
\makeatother

\begin{document}

%\title{Negative as Positive: A Graph Contrastive Learning Method for  Out-of-distribution Generalization}
\title{Negative as Positive: Enhancing  Out-of-distribution Generalization for Graph Contrastive Learning}
\author{Zixu Wang}
\orcid{0009-0006-1327-6366}
\affiliation{%
  \institution{CAS Key Laboratory of AI Safety, Institute of Computing Technology, Chinese Academy of Sciences}
  \department{University of Chinese Academy of Sciences}
  \city{Beijing}
  \country{China}
}
\email{wangzixu22s@ict.ac.cn}

\author{Bingbing Xu}
\orcid{0000-0002-0147-2590}
\affiliation{%
  \institution{CAS Key Laboratory of AI Safety, Institute of Computing Technology, Chinese Academy of Sciences}
  \city{Beijing}
  \country{China}
}
\email{xubingbing@ict.ac.cn}

\author{Yige Yuan}
\orcid{0000-0001-8856-668X}
\affiliation{%
  \institution{CAS Key Laboratory of AI Safety, Institute of Computing Technology, Chinese Academy of Sciences}
  \department{University of Chinese Academy of Sciences}
  \city{Beijing}
  \country{China}
}
\email{yuanyige20z@ict.ac.cn}

\author{Huawei Shen}
\orcid{0000-0003-2425-1499}
\affiliation{%
  \institution{CAS Key Laboratory of AI Safety, Institute of Computing Technology, Chinese Academy of Sciences}
  \city{Beijing}
  \country{China}
}
\email{shenhuawei@ict.ac.cn}

\author{Xueqi Cheng}
\orcid{0000-0002-5201-8195}
\affiliation{%
 \institution{CAS Key Laboratory of AI Safety, Institute of Computing Technology, Chinese Academy of Sciences}
  \city{Beijing}
  \country{China}
}
\email{cxq@ict.ac.cn}


% \renewcommand{\shortauthors}{Wang and Xu, et al.}

\begin{abstract}
  % Graph contrastive learning with supervised fine-tuning is the prevailing paradigm in current graph pre-training， which has achieved significant advancements. However, it's out-of-distribution (OOD) generalization capability has received limited research attention. In this work, we point out that cross-domain negative samples during graph contrastive learning make the distribution gap between different domains larger and draw back the model's OOD generalization performance. To solve this problem, we treat the most similar cross-domain negative samples as positive samples during graph contrastive learning. Experimental results on a multitude of datasets demonstrate that our approach significantly enhances the OOD generalization capability of traditional graph contrastive learning. 
  %In this work, we point out that the traditional composition of \yyg{\sout{positive and negative}} sample pairs in GCL restricts the cross-domain pairs only to be negative, 

  Graph contrastive learning  (GCL), standing as the dominant paradigm in the realm of graph pre-training, has yielded considerable progress. 
  Nonetheless, its capacity for out-of-distribution (OOD) generalization has been relatively underexplored.
  In this work, we point out that the traditional optimization of InfoNCE in GCL restricts the cross-domain pairs only to be negative samples, 
  which inevitably enlarges the distribution gap between different domains. This violates the requirement of domain invariance under OOD scenario and consequently impairs the model's OOD generalization performance.
  To address this issue, we propose a novel strategy ``Negative as Positive'', where the most semantically similar cross-domain negative pairs are treated as positive during GCL. 
  Our experimental results, spanning a wide array of datasets, confirm that this method substantially improves the OOD generalization performance of GCL.
  
\end{abstract}







\maketitle

\section{Introduction}

%XBB：第二大句在描述GCL怎么做的，不需要在引言的一开头写这个，缺点写的很模糊，关注在启发式的定义各种增强，任务loss上，不关注OOD，这个缺点不够本质
%

%XBB： 换一个写作逻辑，先写GCL很重要，GCL的，现有方法，缺点
%第二段写我们剖析这个缺点的原因，PDD是OOD泛化的一个影响factor，我们基于PDD分析，进而往下发现原因，
%基于此分析，我们提出NP算法

Graph Contrastive Learning (GCL) with supervised fine-tuning has emerged as the dominant paradigm for graph pre-training, exhibiting remarkable performance across diverse downstream tasks while requiring only a limited amount of labeled data\cite{GRACE, GCA, Graphcl,BGRL, CSSL,GCC, HeCo, JOAO, DACL, Ginfoclust, Mirat}. Generally, GCL aims at training a graph encoder that maximizes the mutual information between instances with similar semantic information via augmentation.

\begin{figure}[t]
    \centering
    \subfigure{
        \includegraphics[width=0.46\linewidth]{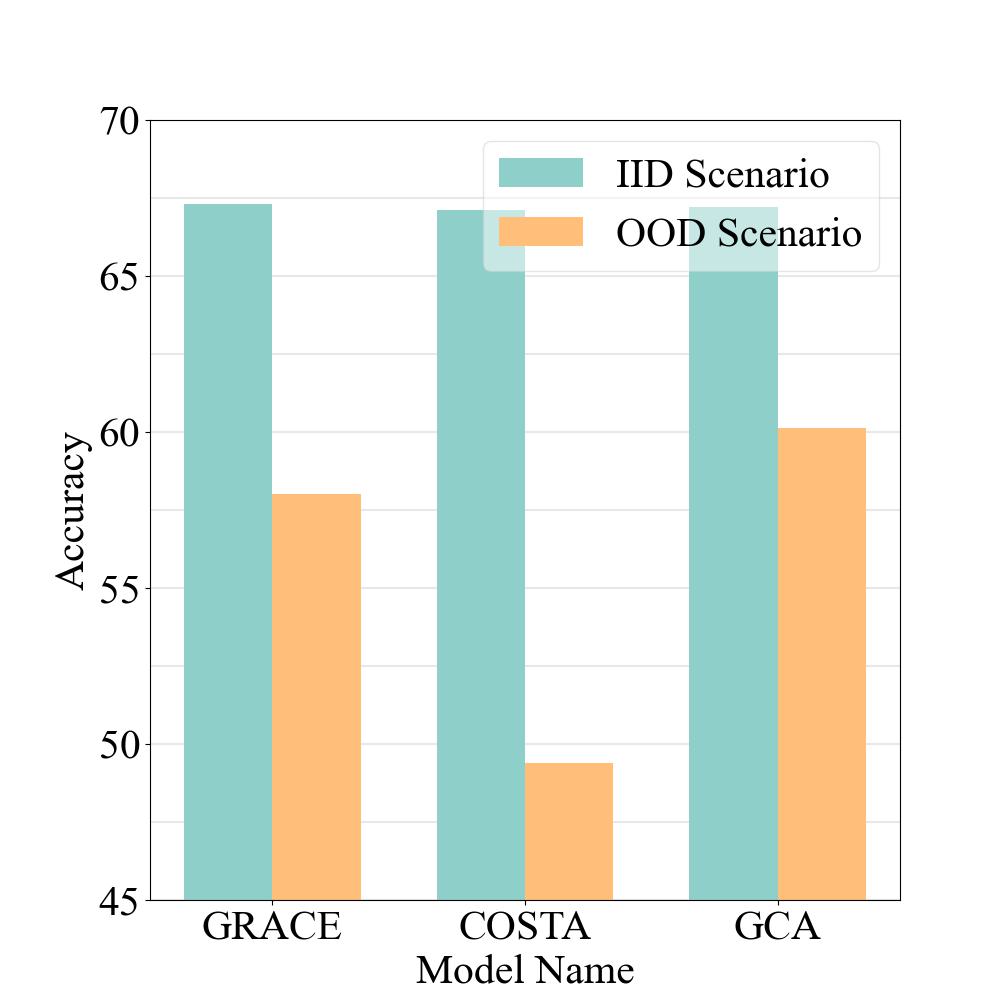}
    }
    \subfigure{
        \includegraphics[width=0.46\linewidth]{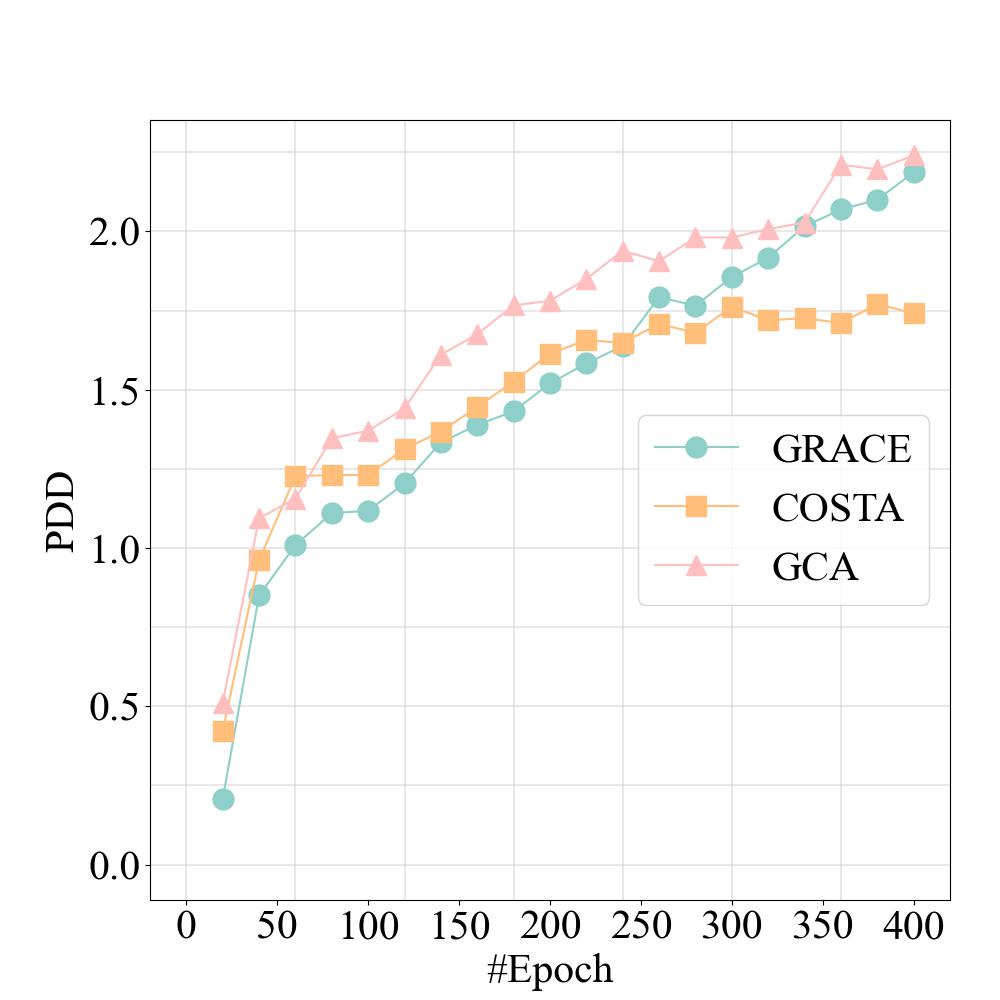}
    }
    \caption{\textbf{Left: Traditional GCLs perform badly under OOD scenario compared to IID one. Right: Pairwize-Domain-Discrepancy grows during GCL.} }
    \label{fig:valid_experiments}
\end{figure}

% 语序，data提前
Most existing works assume the pre-text graph and downstream graph are independent and identically distributed (IID)\cite{GRACE, GCA}.
% Most existing works depend on the independent and identically distributed (IID) assumption between pre-text and downstream dataset. 
However, the graph in the downstream task often exhibits an out-of-distribution (OOD) pattern compared to that encountered in pre-text task\cite{ding2021closer, EERM, GIL, StableGL, DIDA, GSAT, DIR, chen2023causality}.
Furthermore, we find that current methods perform poorly on the OOD downstream graph than IID ones, as shown on the left side of ~\cref{fig:valid_experiments}.

To delve into the phenomenon mentioned above, we utilize pairwise domain discrepancy (PDD), which is widely used in prior works\cite{DFDG, DICA, MDA, CIDG} to measure the model's OOD generalization capability. PDD describes the average distance between domain centers in the embedding space.
% To delve into the phenomenon mentioned above, we utilize pairwise-domain discrepancy (PDD), which is closely related to the model's OOD generalization capability[refs]. PDD measures the average distance between domain centers in the embedding space, and prior works [refs] utilize the minimization of PDD to improve generalization. 
As shown on the right side of ~\cref{fig:valid_experiments}, PDD gradually increases during GCL training, aligning with the declined performance under the OOD scenario. Through in-depth analysis (details in Sec.~\ref{sec:motivation}), we argue that the model's reduced generalization capability stems from treating cross-domain pair as a negative sample solely in the traditional GCL paradigm. By aiming to reduce negative sample similarity in InfoNCE\cite{InfoNCE}, domains are pushed further apart, resulting in increased PDD and poor OOD generalization performance.

Motivated by the above analysis, we propose \textbf{N}egative \textbf{a}s \textbf{P}ositive, namely \textbf{NaP}, to enhance the OOD generalization of GCL.
Specifically, considering that the embedding of nodes represents its semantics, NaP dynamically transfers a subset of cross-domain negative samples as positive samples based on the embedding similarity, and reduces the distance of positive samples.
Therefore, NaP can narrow the distribution gap among embedding from different domains, further preserving domain-shared knowledge and enhancing OOD generalization.
Extensive experiments on various datasets and tasks demonstrate the improved domain generalization capability of the proposed method compared to the SOTA GCL methods.

% 1. 只保留theta; 
% 2. 待定；

\section{Preliminaries}
% In this section, we first give the notations and the paradigm of GCL with supervised fine-tuning, then introduce the definition of Out-of-Distribution in GCL.

\subsection{Task Formulation of OOD in GCL}
%定义图
Let $\mathcal{G}=(\mathbf{X, A})$ denote a graph, where $\mathbf{X}\in \mathbb{R}^{N\times F}$ denotes the nodes' feature map, and $\mathbf{x}_i$ is the feature of node $v_i$. $\mathbf{A} \in \mathbb{R}^{N \times N}$ denotes the adjacency matrix, where $\mathbf{A}_{ij} = 1$ means $v_i$ and $v_j$ are connected. 
%\sout{⟺(vi,vj)∈E\iff (v_i, v_j) \in \mathcal{E}}. 
%定义GCL
As Eq.~\ref{equ: min_im} shows, GCL aims at training a GNN encoder\cite{SGC, GCN, GAT, GIN} $g_{\theta}(\mathcal{G})$ by maximizing the mutual information between instances with similar semantic information via augmentation. The augmented graph is noted as ${\mathcal{G}}_{\psi}$, where $\psi$ represents one kind of augmentation method such as used in \cite{GRAND, GRACE, GCA, graphmae},  
% XBB：这儿是用INfoNCE 的原始形式，还是MI
%\tilde
\begin{equation}
    \label{equ: min_im} 
    \theta^* = \max_{\theta} \mathcal{I}(g_{\theta}({\mathcal{G}_{\alpha}}), g_{\theta}({\mathcal{G}_{\beta}})) 
\end{equation}

The formulation of OOD in GCL is as follows: $\theta^*$ in Eq.~\ref{equ: min_im} is optimized on data $\{(G^i)|_{i=1}^S\}$, and leveraged to infer $G^T$, with $P(G^T) \neq P((G^i)|_{i=1}^S)$, where $S$ is the number of domains in pre-training. 
In contrast, within IID scenarios, $P(G^T) = P((G^i)|_{i=1}^S)$.
Fig.\ref{fig:valid_experiments} shows the test accuracy for OOD and IID scenarios of a representative benchmark GOOD-Twitch, where each graph $G^i$ is a gamer network and different domains represent the different languages used in the network. %这个接的应该是GOODTwitch的domain的划分方法
All three GCL methods\cite{GRACE, costa, GCA} exhibit significant performance degradation in the presence of OOD, emphasizing the critical importance of investigating this phenomenon.

\subsection{Pairwise Domain Discrepancy}
Pairwise domain discrepancy(PDD) is widely used to measure the model’s OOD generalization capability in prior works\cite{DFDG, DICA, MDA, CIDG}.
It's the average distance among all pairs of the domains' centers.
Denote the center embedding of domain $d$ as $\bar{h^d} = \frac{1}{N_d}\sum_{i=1}^{N_d}\mathbf{H}_i^d$, and PDD is as follows:
\begin{equation}
    PDD = \frac{1}{\binom{P}{2}} \sum_{p,q|_{1 \leq p < q \leq P}}\|\bar{h^p} - \bar{h^q}\|,
\end{equation}
where P denotes the number of domains, $\mathbf{H}_i^d$ denotes the embedding of $i$-th node in domain $d$ and ${N_d}$ denotes the number of nodes in domain $d$.

\section{Proposed Method}
% \begin{figure}[t]
%     \centering
%     \includegraphics[width=0.8\linewidth]{figs/fig3.jpg}
%     \caption{\textbf{Schematic diagram of CDNs' effect. The blue arrows are pushing the CDNs away while the red reversed-arrows are pulling positive samples near.} }
%     \label{fig:effect}
% \end{figure}
\begin{figure}[t]
    \centering
    \subfigure{
        \includegraphics[width=0.46\linewidth]{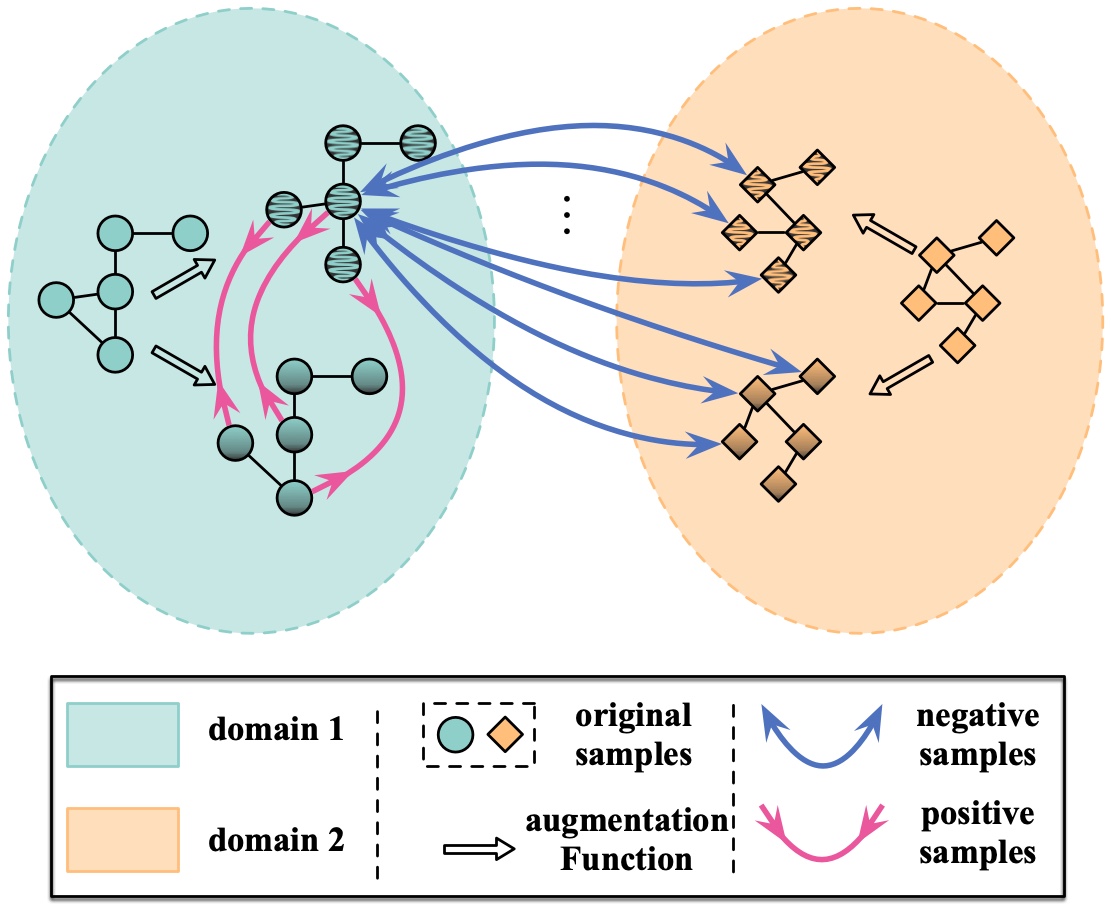}
    }
    \subfigure{
        \includegraphics[width=0.46\linewidth]{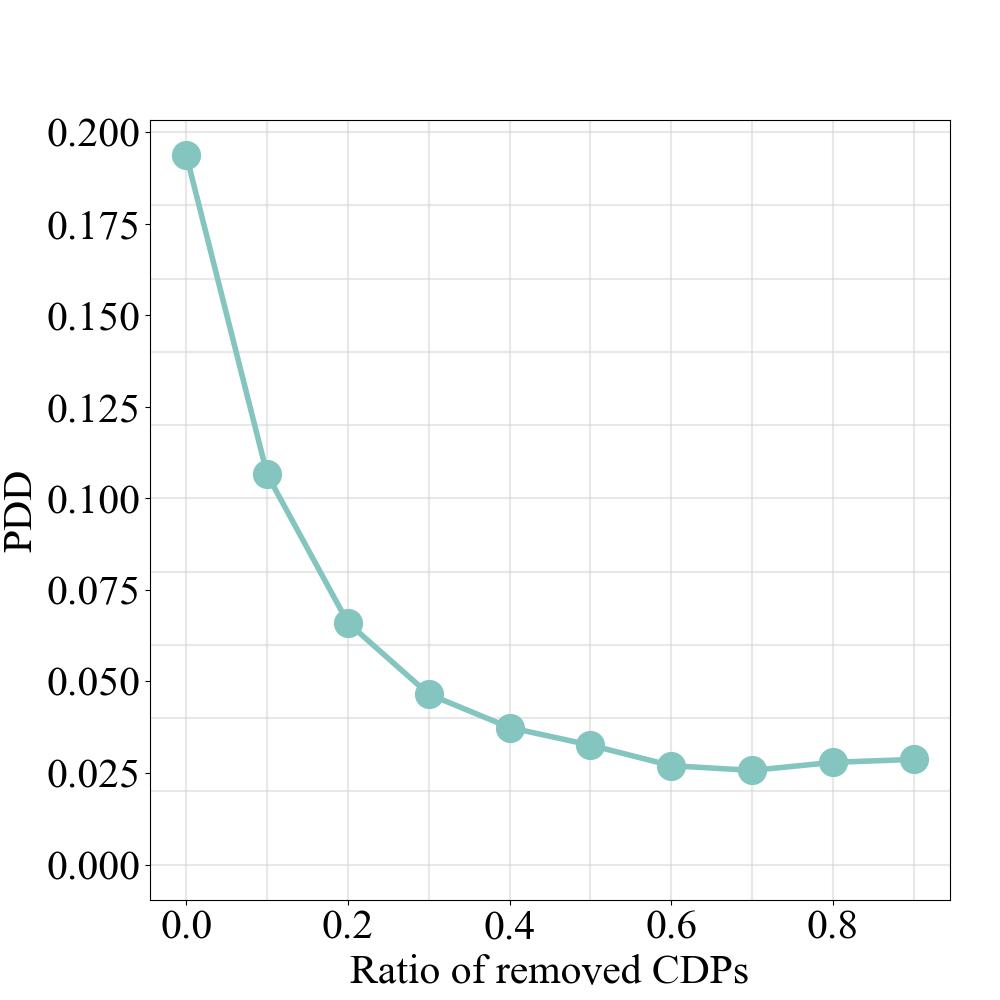}
    }
    \caption{\textbf{Left: All CDPs are negative samples. Right: PDD decreases while more CDPs are removed.}}
    \label{fig:motivation}
\end{figure}
In this section, we first show the motivation of NaP and then introduce each part of NaP in detail.

\subsection{Motivation}\label{sec:motivation}
% In reality, the phenomenon of OOD is highly prevalent in GCL.
% As is illustrated by two common scenarios: in social networks, GCL may be trained on highly influential communities but applied to low-influence users [refs]. Similarly, in investment networks, GCL might be trained on numerous high-market-value companies but applied to small and medium-sized enterprises. 
% Such examples highlight the critical need to address OOD issues in GCL.

The phenomenon of OOD is highly prevalent in GCL, which underscores the need to address OOD issues.
Taking one common scenario as an example: in social networks, GCL may be trained on highly influential communities but applied to low-influence users \cite{silent_majority}. 
%Similarly, \yyg{in investment networks,} GCL might be trained on numerous high-market-value companies but applied to small and medium-sized enterprises\yyg{\sout{in investment networks [refs]}}. 
This phenomenon is also common in areas such as financial risk prediction\cite{bridge_gnn} (high-market-value companies VS medium-sized ones) and fraudulent accounts detection (old fraudulent style VS new ones).
% \yyg{\sout{Such commonality underscores the need to address OOD issues.}}
Such commonality highlights the critical need to address OOD in GCL.
% \yyg{\sout{We have validated that the traditional GCLs perform badly at OOD target and the PDD of all domains keeps increasing during GCL's training as shown in fig.  }}
However, as shown in~\cref{fig:valid_experiments}, the traditional GCLs perform poorly on OOD scenarios, and the PDD of all domains continues to increase during the training of GCLs.
The increasing PDD indicates that GCL will widen the gap in domain distribution and push domains further apart, 
violating an ideal OOD generalization, which should capture the shared knowledge among different domains and facilitate the seamless transfer to unseen target domains.
%However, it is undesirable because an effective OOD-generalized model should capture the shared knowledge among source domains, facilitating the seamless transfer of learned embedding to unseen target domains.

Let \textbf{C}ross-\textbf{D}omain \textbf{P}air (CDP) represent two nodes from different domains. We argue that the principal constituents of negative samples for optimizing Eq. ~\ref{equ: min_im} are CDPs, being a significant factor in the poor OOD generalization capability.
Specifically, as shown on the left side of Fig.\ref{fig:motivation}, CDPs can only be negative samples, and the traditional contrastive loss will decrease the similarity of negative samples, leading to the pushing-apart effect between the nodes in CDP.
Furthermore, as shown on the right side of Fig.~\ref{fig:motivation}, the PDD of node embedding of GCL decreases as the ratio of removed CDP increases which proves that CDPs are harmful to GCL's OOD generalization. 
% 加入 pdd-rv cross negative samples;  再跑一个domain内的平均距离（如果domain 内的不变，则可以不写message passing）；
% \sout{Furthermore, due to the weaker impact of message-passing mechanisms between different domains compared to within-domain interactions, samples within-domain tend to be closer than CDPs.}
Therefore, the CDPs in traditional GCL tend to push the representations of samples from different domains apart, resulting in a higher PDD and a poor OOD generalization ability.

% \begin{figure}[htbp]
%     \centering
%     \includegraphics[width=0.60\linewidth]{figs/PDD_Ratio.jpg}
%     \caption{\textbf{The impact of CDPs on PDD. }}
%     \label{fig:pdd_ratio}
% \end{figure}

\subsection{NaP: Negative as Positive}\label{sec:Method}
% 和infonce 耦合； 和GRACE解绑；
% 在实验部分提出follow GRACE 进行了实现；

\begin{figure}[htbp]
    \centering
    \includegraphics[width=\linewidth]{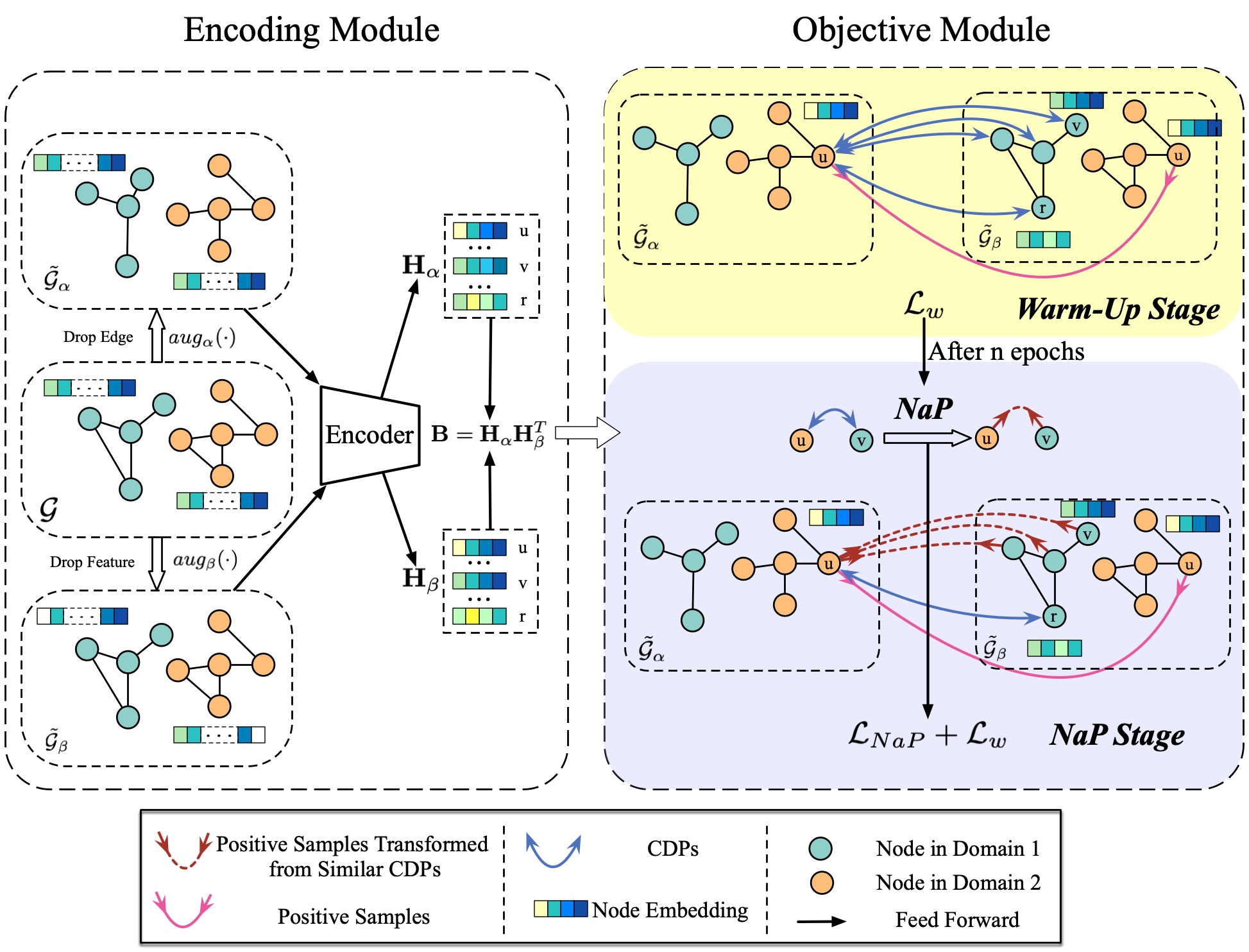}
    \caption{\textbf{The overall framework of NaP consists of two modules: the encoding module and the objective module. The objective module comprises two stages: the warm-up stage and the NaP stage. }}
    %After several rounds of training in the warm-up stage, we proceed to the NaP stage.} }
    \label{fig:framework}
\end{figure}

% Encoding + Objective Module;
% Encoding, (warm up stage + NaP stage)
Based on the above motivation, we propose NaP, which transfers a subset of the most semantically similar negative samples as positive ones. 
Fig.\ref{fig:framework} illustrates the overall framework of NaP, including the encoding module and the objective module. 
%In the objective module, there are two stages: warm-up stage and NaP stage. 
Note that our NaP framework can be adapted to existing GCL methods that use InfoNCE as loss function, e.g., GRACE\cite{GRACE}, GCA\cite{GCA}, and GraphCL\cite{Graphcl}. 

\subsubsection{Encoding Module}
%XBB:先讲这个part的目的
The objective of this module is to obtain the embedding of each node. 
We first generate different views of $\mathcal{G}$ as $\tilde{\mathcal{G}_{\alpha}}$, $\tilde{\mathcal{G}_{\beta}}$ using graph augmentations. And input the augmented graphs into a shared GCN\cite{GCN} encoder to get the embedding $\mathbf{H}_{\alpha}$, $\mathbf{H}_{\beta}$. The propagation of the $l$-th layer of GCN is represented as: 
\begin{equation}
    \mathbf{H}^{l+1} = \sigma( \mathbf{\tilde{D}}^{-\frac{1}{2}} \mathbf{\tilde{A}} \mathbf{\tilde{D}}^{-\frac{1}{2}} \textbf{H}^{l} \textbf{W}^{l}),
\end{equation}
where $\sigma(\cdot)$ is the activation function, $\mathbf{\tilde{A}}$ is the adjacency matrix with self-loop, $\mathbf{\tilde{D}}$ is the corresponding degree matrix and $\textbf{W}$ is the parameter matrix.

\subsubsection{Objective Module} 
Considering that the representations obtained from randomly initialized models may not accurately reflect the semantic information of the samples, we have to train the GCL in the traditional way for several epochs. Therefore, there are two stages in this module: Warm-up stage and NaP stage.

\paragraph{(1) Warm-Up Stage:}
Firstly, we use the traditional InfoNCE loss to train the GCL as the warm-up for the NaP stage.
The InfoNCE loss for each positive pair $(v_{\alpha i}, v_{\beta i})$ in warm-up stage is:
% \begin{equation}
%     \label{equ:warm-up_loss}
%     l_{w} = 
%     -\log {
%     \frac{e^{\frac{\theta(v_{\alpha i}, v_{\beta i})}{\tau}}}
%         { 
%         e^{\frac{\theta(v_{\alpha i}, v_{\beta i})}{\tau}} + 
%         \sum_{j\neq i}e^{\frac{\theta(v_{\alpha i}, v_{\beta j})} {\tau} } + \sum_{j\neq i}e^{\frac{\theta(v_{\alpha i}, v_{\alpha j})}{\tau}}
%         }
%     }
% \end{equation}
\begin{equation}
\footnotesize
    \label{equ:warm-up_loss}
    \mathcal{L}_{w} = 
    -\log {
    \frac{\exp(\frac{\theta(v_{\alpha i}, v_{\beta i})}{\tau})}
        { 
        \exp(\frac{\theta(v_{\alpha i}, v_{\beta i})}{\tau}) + 
        \sum_{j\neq i} \exp({\frac{\theta(v_{\alpha i}, v_{\beta j})} {\tau} }) + \sum_{j\neq i}\exp({\frac{\theta(v_{\alpha i}, v_{\alpha j})}{\tau}})
        }
    }
\end{equation}
The $\theta(v_{\alpha i}, v_{\beta j})$ means cosine similarity between $\mathbf{H}_{\alpha i}$, $\mathbf{H}_{\beta j}$.
% The two views are symmetric and the overall contrastive loss can be written as 
% \begin{equation}
% \label{equ:GRACE_loss_overall}
%     \mathcal{L}_{GRACE} = \frac{1}{2N}\sum_{i=1}^N[l_{G}(v_{\alpha i}, v_{\beta i}) + l_{G}(v_{\beta i}, v_{\alpha i})]
% \end{equation}

%如上面讲的，Loss中的负样本是导致效果差的主要原因，我们提出将部分负样本转化为正样本，通过拉近正样本之间的距离进而使得不同domain的节点靠拢

\paragraph{(2) NaP Stage:}
% After n epochs warm-up,  we enter the NaP stage. To mitigate the domain discrepancies introduced by CDPs, we choose a subset of CDPs to transform into positive samples. 
After n epochs warm-up, we enter the NaP stage where a subset of CDPs is chosen to transform into positive samples to mitigate the domain discrepancies introduced by CDPs.
We select the most similar CDPs based on the between-view embedding similarity in the current epoch and transform the chosen CDPs into positive samples by adding a new loss item. Firstly, we compute the between-view-similarity matrix: 
\begin{equation}
    \mathbf{B} = \mathbf{H}_{\alpha} \mathbf{H}_{\beta}^T 
\end{equation}
We focus our attention on cross-domain samples, so we update $\mathbf{B}$ as follows:
\begin{equation}
    \mathbf{B}_{ij} = 0 \text{ if } d_i = d_j 
\end{equation}

% IDX : 公式化；换个表示方式；

The $d_i$ means the domain index of $v_i, i \in \{1, 2, ..., N\}$. After sorting the elements in $\mathbf{B}$, we can select the top $r$ of most similar samples and their indices $idx$ as follows:
\begin{equation}
    \mathrm{idx} = \arg \max_{I \subset \mathbb{R}^{N \times N}: |I| = r } \sum_{(i,j) \in I} \mathbf{B}_{ij}
\end{equation}
To obtain the transformed CDPs, we set the mask matrix: 
\begin{equation}
    mask_{ij} = 1 \text{ if } (i,j) \in \mathrm{idx} \text{ else } 0
\end{equation}
Up to this point, only the top $r$ most similar CDPs are retained in the mask. We add a new loss item to transform these CDPs into positive samples, namely $\mathcal{L}_{NaP}$:
% \begin{equation}
%     \label{equ:NaP_loss}
%    l_n = -\log {
%         \frac{
%             \sum_{j\neq i}
%             mask_{ij} 
%             \{
%                 e^{\frac{\theta(v_{\alpha i}, v_{\beta j})}{\tau}}+ 
%                 e^{\frac{\theta(v_{\alpha i}, v_{\alpha j})}{\tau}} 
%             \}
%         }
%         { 
%             e^{\frac{\theta(v_{\alpha i}, v_{\beta i})}{\tau}} + 
%             \sum_{j\neq i}e^{\frac{\theta(v_{\alpha i}, v_{\beta j})}{\tau}} + 
%             \sum_{j\neq i}e^{\frac{\theta(v_{\alpha i}, v_{\alpha j})}{\tau}} 
%         }
%     }  + l_w 
% \end{equation}
\begin{equation}
    \footnotesize
    \label{equ:NaP_loss}
    \mathcal{L}_{NaP} = -\log {
            \frac{
                \sum_{j\neq i}
                mask_{ij} 
                \{
                    \exp({\frac{\theta(v_{\alpha i}, v_{\beta j})}{\tau}})+ 
                    \exp({\frac{\theta(v_{\alpha i}, v_{\alpha j})}{\tau}}) 
                \}
            }
            { 
                \exp({\frac{\theta(v_{\alpha i}, v_{\beta i})}{\tau}}) + 
                \sum_{j\neq i}\exp({\frac{\theta(v_{\alpha i}, v_{\beta j})}{\tau}}) + 
                \sum_{j\neq i}\exp({\frac{\theta(v_{\alpha i}, v_{\alpha j})}{\tau}}) 
            }
        }
\end{equation}
Finally, for each positive pair $(v_{\alpha i}, v_{\beta i})$, the loss in NaP stage is written as below: 
\begin{equation}
\footnotesize
\label{equ:Overall_loss}
    \begin{split}
        \mathcal{L} &= \mathcal{L}_{NaP} + \mathcal{L}_w \\
                    &= -\log {
                        \frac{\exp(\frac{\theta(v_{\alpha i}, v_{\beta i})}{\tau}) + \sum_{j\neq i}
                                mask_{ij} 
                                \{
                                    \exp({\frac{\theta(v_{\alpha i}, v_{\beta j})}{\tau}})+ 
                                    \exp({\frac{\theta(v_{\alpha i}, v_{\alpha j})}{\tau}}) 
                                \}}
                            { 
                            \exp(\frac{\theta(v_{\alpha i}, v_{\beta i})}{\tau}) + 
                            \sum_{j\neq i} \exp({\frac{\theta(v_{\alpha i}, v_{\beta j})} {\tau} }) + \sum_{j\neq i}\exp({\frac{\theta(v_{\alpha i}, v_{\alpha j})}{\tau}})
                            }
                        }
    \end{split}
\end{equation}

% To this end, the loss for one positive sample in NaP stage is: 
% \begin{equation}
%     l_{N} = l_{1} + l_{2}
%     \label{equ:NaP_loss_one_pair}
% \end{equation}

% And the total loss is just like Eq. but exchange lGl_{G} with lNl_{N}: 
% \begin{equation}
%     \label{equ:NaP_loss_overall}
%     \mathcal{L}_{NaP} = \frac{1}{2N}\sum_{i=1}^N[l_{N}(v_{\alpha i}, v_{\beta i}) + l_{N}(v_{\beta i}, v_{\alpha i})]
% \end{equation}

To sum up, after n epochs of training according to the loss in Eq.~\ref{equ:warm-up_loss}, NaP selects the top r most similar CDPs based on the current epoch's embedding similarity. These CDPs are then treated as positive samples, and the training continues using the loss described in Eq.~\ref{equ:Overall_loss}.

\section{Experiments}
In this section, we empirically evaluate the quality of produced node embedding on node classification using two public benchmark datasets: GOOD benchmark and Facebook100.

\subsection{Datasets}
We use 3 datasets from GOOD benchmark\cite{GOOD} and 15 datasets from Facebook100\cite{FB100} for experiments. Datasets from Facebook100 are social networks of 100 universities in the US. Each university is viewed as a domain and each node stands for a student or faculty. 

\begin{table}[htbp]
    \small
    \centering
    \caption{\textbf{Experiments results of all baselines and NaP. The bold font represents the top-1 performance and the underline represents the second performance across the self-supervised methods. }}
    \label{tab:ood_res}
    \begin{adjustbox}{width=\linewidth}
    \setlength{\tabcolsep}{0.1mm}
    \begin{tabular}{c|ccccc|ccc}
        \toprule
         & \multicolumn{5}{|c|}{\textbf{Facebook100} } & \multicolumn{3}{|c}{\textbf{GOOD benchmark} } \\
        Dataset & Santa & Wake & Bucknell & Colgate & Wesleyan & Twitch & CBAS & Cora\\
        Domain &  \multicolumn{5}{|c|}{university } &  language &  color & degree\\
        \midrule
        DGI & 87.08\% & 83.02\% & 89.24\% & 89.55\% & 88.52\% & 53.34\% & \underline{52.86\%}& 46.61\%\\
        GRACE & 87.88\% & 82.70\% & 90.12\% & 82.09\% &\underline{90.80\%}& 58.00\% &48.10\% & 50.85\%\\
        GCA & 89.10\% & 82.71\% & \underline{93.01}\% & 91.18\% & 90.11\% & 60.14\%& 50.00\% & \underline{50.97\%}\\
        COSTA & 89.93\% & 75.29\% & 91.46\% & \underline{91.52}\% & 88.36\% & 49.40\% & 45.24\% & 48.09\%\\
        BGRL & 88.80\% & \underline{83.61}\% & 91.59\% & 85.18\% & 82.41\% & \textbf{63.25\%}& 49.05\%& 40.63\%\\
        MVGRL & \underline{90.12}\% & 78.58\% & 91.45\% & 88.38\% & 90.13\% & 53.98\% & 50.95\%& 47.15\%\\
        \midrule
        Ours & \textbf{91.06\%} & \textbf{86.55\%} & \textbf{93.26}\%  & \textbf{93.18\%} & \textbf{91.51\%} & \underline{61.08\%} & \textbf{53.33\%} &  \textbf{51.31\%}\\
        improve & +0.94\% & +2.94\% & +0.25\% & +1.66\% & +0.71\% & -2.17\% & +0.47\% & +0.34\% \\
        \midrule
        GCN & 92.10\% & 87.14\% & 94.47\% & 93.24\% & 92.10\% & 51.65\% & 65.24\% & 59.39\% \\
        
        \bottomrule
    \end{tabular}
    \end{adjustbox}
\end{table}

\subsection{Experimental Setup}
\subsubsection{Data settings}
We divide the dataset according to GOOD\cite{GOOD}. Specifically, for the Facebook100, we randomly use 9 domains as the source domains for training, 1 domain (Emory) for validation, and 15 others for testing.

\subsubsection{Model and Metric settings}
We use 6 contrastive methods: DGI, GRACE, GCA, COSTA, BGRL, MVGRL\cite{DGI, GRACE, GCA, BGRL, costa, MVGRL} for self-supervised methods, and use GCN\cite{GCN} as supervised baselines. The checkpoint for OOD testing is decided based on the result obtained from OOD validation domains. The reported results represent the average accuracy from three independent runs.

\subsubsection{Results and Analysis}

\paragraph{(1) NaP surpasses baselines}
As shown in the Table.\ref{tab:ood_res}, NaP outperforms almost all GCL baselines. It is worth noting that NaP surpasses all four baselines - DGI, GRACE, GCA and COSTA\cite{DGI,GRACE, GCA, costa} - that use InfoNCE loss, with an improvement of up to 11.68\%. Furthermore, NaP outperforms BGRL, which uses BYOL\cite{BGRL} as the loss function, and MVGRL, which uses JSD\cite{MVGRL} as the loss function, on the majority of datasets. 
Last but not least, compared to GCN\cite{GCN}, NaP has a relatively good performance considering we use significantly fewer labels. 
% \sout{We observe a more significant improvement of NaP on the Facebook100 compared to GOOD in Table. ref{tab:ood_res}. We speculate that this is due to the presence of inter-domain edges in GOOD, while there are no such edges in Facebook100. Under the influence of message passing, the differences between domains in the GOOD are smaller which overlaps with the target of NaP and leads to a smaller degree of improvement in GOOD.}

\begin{figure}[htbp]
    \centering
    \includegraphics[width=0.85\linewidth]{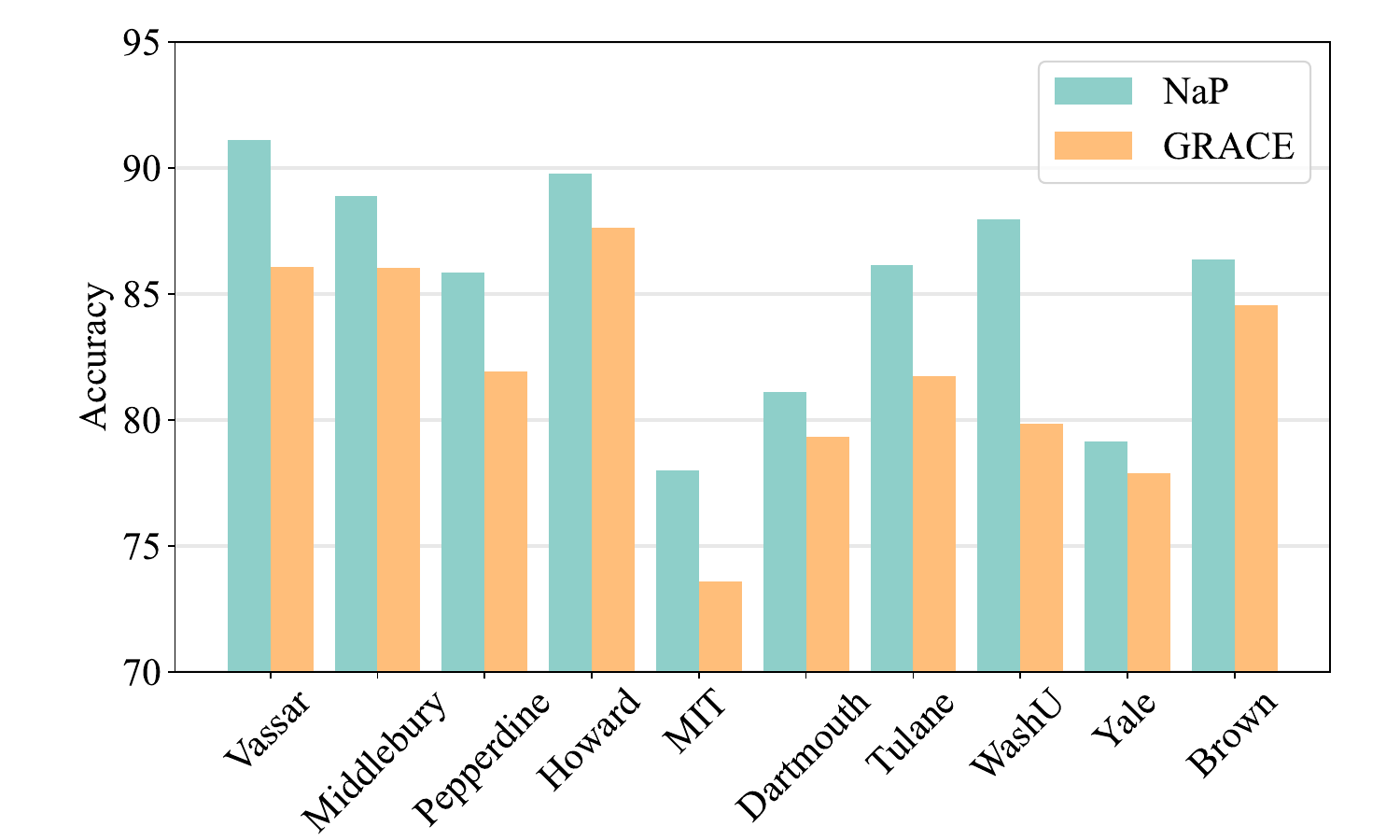}
    \caption{\textbf{Experiments result of NaP and GRACE on 10 OOD target domains from Facebook100. }}
    \label{fig:fb_compare}
\end{figure}

\paragraph{(2) NaP's strategy is highly effective.}
As shown in Fig.\ref{fig:fb_compare}, NaP achieves higher accuracy on 10 additional domains. Since this experiment utilized GRACE as a warm-up stage, NaP's superior OOD generalization ability demonstrates the effectiveness of the proposed strategy in this paper.

\begin{figure}[htpb]
    \centering
    \includegraphics[width=0.85\linewidth]{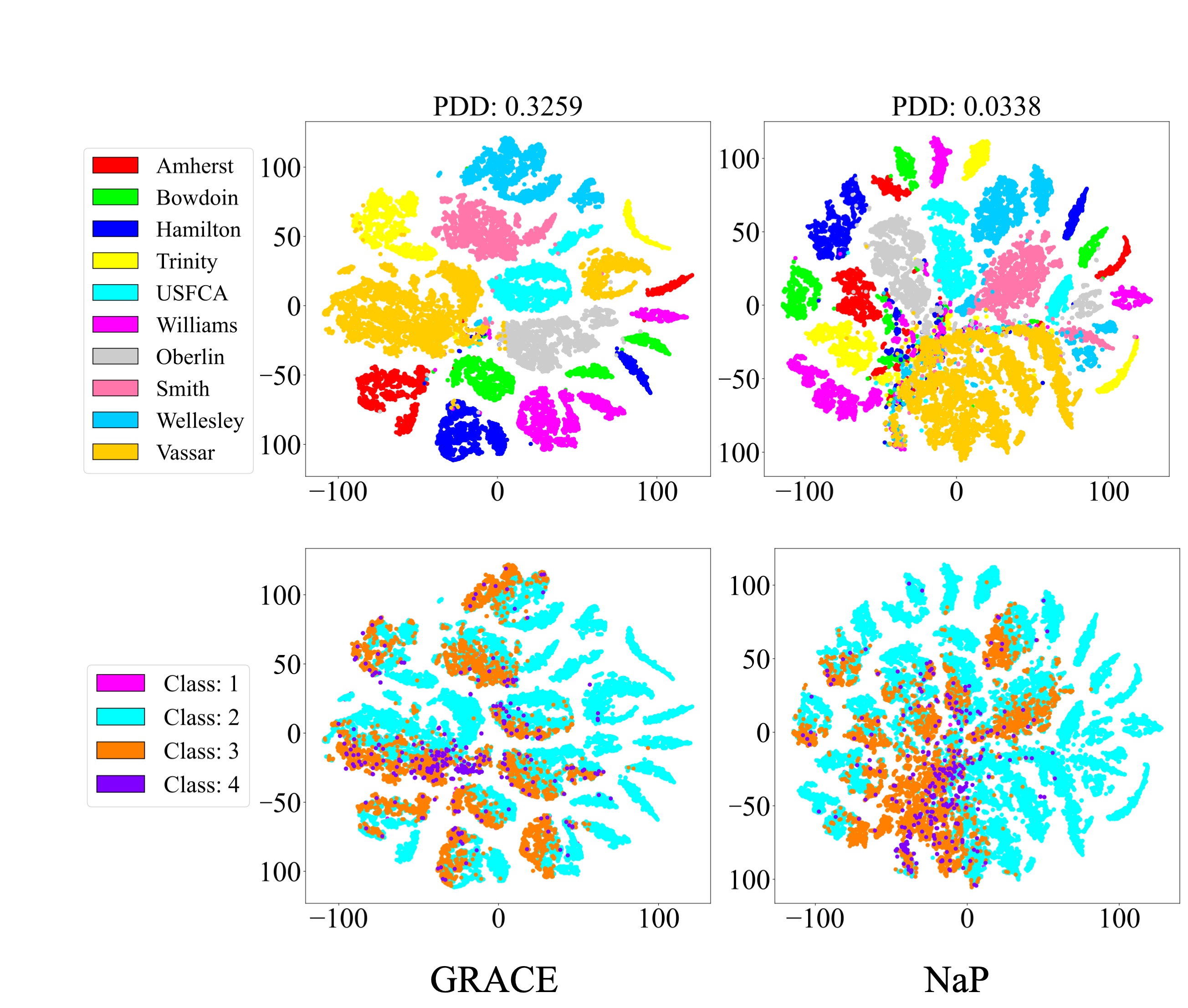}
    \caption{\textbf{t-SNE visualization and PDD of node embedding.}}
    \label{fig:node_emb}
\end{figure}

\paragraph{(3) NaP narrows the distance between domains}
As shown in Fig.~\ref{fig:node_emb}, compared to GRACE, the embedding obtained by NaP exhibits a smaller PDD.  More importantly, as the PDD decreases, the node distributions between different domains with the same label become closer.

\begin{table}[htbp]
    \centering
    \small
    \caption{\textbf{The similarity comparison of different CDPs.}}%\textbf{Mean similarity compare between input space and embedding space.}}
    \begin{tabular}{c|cc}
         & Input Feature & Embedding \\
       \midrule
      All CDPs & 0.0015 & 0.0199\\
      Transformed CDPs & 0.0282 & 0.8523\\
      Other CDPs& -0.0010 & -0.0891\\
      \bottomrule
    \end{tabular}
    \label{tab:sim_compare}
\end{table}

\paragraph{(4) The CDPs transformed by NaP exhibit semantic similarity in the input space.}
As shown in Table.\ref{tab:sim_compare} the cosine similarity of all transformed CDPs is significantly higher than that of all CDPs and the remaining CDPs. This demonstrates that NaP indeed transforms the most semantically similar CDPs into positive samples.

% \subsubsection{NaP transforms the most semantically similar CDPs into positive samples}
% We compared the feature similarity in the input space with the representation similarity obtained through NaP as shown in the following table. It can be observed that overall, NaP improves the similarity between domains. Additionally, it is worth noting that, compared to the overall average, the CDPs selected by NaP for conversion into positive samples exhibit higher similarity. This indicates that NaP effectively transforms some semantically similar CDPs in the input space into positive samples.

\section{Conclusion}
In this work, we investigate the OOD generalization capability of traditional graph contrastive learning methods. We argue that cross-domain pairs (CDPs) make the domains distribution shift larger and hinder the model's OOD generalization capability. Based on this, we propose to transfer the most semantically similar CDPs as positive samples. Comprehensive experiments show that our method NaP significantly benefits the OOD generalization capability of graph contrastive learning methods.

\section{Acknowledgement}
This work was supported by the National Natural Science Foundation of China (Grant No.U21B2046, No.62202448), the Strategic Priority Research Program of the CAS under Grants No. XDB0680302.

\bibliographystyle{ACM-Reference-Format}
\balance
% \bibliography{refs}

\end{document}